\documentclass[sigconf]{acmart}

\AtBeginDocument{%
  }

\setcopyright{rightsretained}
\copyrightyear{2024}
\acmYear{2024}
\acmDOI{}
\acmISBN{}

\acmConference[AIBS '24]{The First Worshop on AI Behavioral Science}{August 25,
  2024}{Barcelona, Spain}

\begin{document}

\title{Adaptive Behavioral AI: Reinforcement Learning to Enhance Pharmacy Services}

\author{Ana Fernández del Río}
\email{ana@causalfoundry.ai}
\affiliation{%
  \institution{Causal Foundry}
  \city{Barcelona}
  \country{Spain}
}

\author{Michael Brennan Leong}
\email{leong@swiperxapp.com}
\affiliation{%
  \institution{SwipeRx}
  \city{Jakarta}
  \country{Indonesia}
}

\author{Paulo Saraiva}
\email{paulo@causalfoundry.ai}
\affiliation{%
  \institution{Causal Foundry}
  \city{Barcelona}
  \country{Spain}
}

\author{Ivan Nazarov}
\email{ivan@causalfoundry.ai}
\affiliation{%
  \institution{Causal Foundry}
  \city{Barcelona}
  \country{Spain}
}

\author{Aditya Rastogi}
\email{aditya@causalfoundry.ai}
\affiliation{%
  \institution{Causal Foundry}
  \city{Barcelona}
  \country{Spain}
}

\author{Moiz Hassan}
\email{moiz@causalfoundry.ai}
\affiliation{%
  \institution{Causal Foundry}
  \city{Barcelona}
  \country{Spain}
}

\author{Dexian Tang}
\email{dexian@causalfoundry.ai}
\affiliation{%
  \institution{Causal Foundry}
  \city{Barcelona}
  \country{Spain}
}

\author{África Periáñez}
\email{africa@causalfoundry.ai}
\affiliation{%
  \institution{Causal Foundry}
  \city{Barcelona}
  \country{Spain}
}

\renewcommand{\shortauthors}{Fernández del Río and Leong, et al.}

\begin{abstract}
Pharmacies are critical in healthcare systems, particularly in low- and middle-income countries. Procuring pharmacists with the right behavioral interventions or nudges can enhance their skills, public health awareness, and pharmacy inventory management, ensuring access to essential medicines that ultimately benefit their patients. We introduce a reinforcement learning operational system to deliver personalized behavioral interventions through mobile health applications. We illustrate its potential by discussing a series of initial experiments run with SwipeRx, an all-in-one app for pharmacists, including B2B e-commerce, in Indonesia. The proposed method has broader applications extending beyond pharmacy operations to optimize healthcare delivery.
\end{abstract}

\ccsdesc[500]{Computing methodologies~Artificial intelligence}
\ccsdesc[500]{Computing methodologies~Sequential decision making}
\ccsdesc[300]{Information systems~Computational advertising}
\ccsdesc[500]{Applied computing~Law, social and behavioral sciences}
\ccsdesc[500]{Applied computing~Online shopping}

\keywords{reinforcement learning, behavioral AI, e-commerce recommendations, adaptive interventions}
\linepenalty=1000

\received{20 February 2007}
\received[revised]{12 March 2009}
\received[accepted]{5 June 2009}

\maketitle


\section{Introduction}

Pharmacies are a crucial element of healthcare delivery everywhere and are often the first line of patient contact. For many patients in low— and middle-income countries (LMICs) who face significant barriers to accessing primary care facilities, pharmacies are the only option for obtaining essential healthcare services such as testing, drugs, and treatment advice. While the role of pharmacies in promoting medication adherence and responsible usage, and providing patient-centered care is unquestionable, pharmacies in many LMICs face substantial challenges, including a shortage of well-trained pharmacists, adequate channels to remain connected to the rest of the healthcare system and updated on emerging public health threats, medication availability and indications, and fragmented and undercapitalized supply chains that complicate adequate restocking and inventory management~\citep{Miller2016, Ikhile2018, Hamid2020}. 

\subsection{Nudges for Pharmacists and Public Health}

Pharmacists, and ultimately their patients, can hence benefit from digital tools and mobile applications that can help them manage their stocks and order supplies, connect to other facilities and peers, refine their skills, and access updated public health news and drug reference guides with the latest information on applicability, clinical requirements, side effect prevalence, contraindications, and pharmacological substitutes. These tools can turn pharmacies from drug procurement points into health hubs where patients can get treatment advice, testing, and health education. 

Equipping these tools with data-driven behavioral nudges contributes to keeping their users adequately engaged. It can provide them with additional timely support as needed, be that in the form of an adaptive continuous professional development program, restocking reminders based on demand prediction, patient-specific prescription recommendations, or actual incentives to reward adherence to medical guidelines (e.g., requiring a positive malaria test before dispensing antimalarials). We propose an Artificial Intelligence (AI) data-centric platform (described in Section \ref{sec:platform}) that can integrate with different tools to enrich them with nudging capabilities through Reinforcement Learning (RL) based recommendations that can be delivered directly through the applications.

We will illustrate this adaptive intervention framework through ongoing experiments with a concrete mobile health solution for pharmacists (described in Section \ref{sec:swiperx}). More technical details and a more thorough discussion of the results, including an additional experiment, can be found in a longer version of this paper \cite{customerjourney}. The same approach, however, can be used with other digital tools within the broader health ecosystem, including applications for patient management \cite{epidamik}, clinical decision support, capacity building, reporting, and efficiency monitoring for patients, healthcare workers, facilities, and decision-makers.  See \cite{perianez2024} for a general discussion in the context of health systems. Substantial transformative potential for health systems lies in such behavioral nudging frameworks.

\subsection{The AI Platform}
\label{sec:platform}

Architecturally, the platform consists of a \emph{Software Development Kit} (SDK), which is embedded into mobile apps and digital tools; analytics, model management, intervention, and adaptive experimentation \emph{frontend} interfaces; and the \emph{backend server}. The backend is the backbone of the platform, responsible for log ingestion, data organization, and labeling, scheduled dispatch of nudges (push notifications, in-app messages, and adaptive UI components), as well as hosting predictive modeling and algorithmic decision-making engine and facilitating configuration thereof through the frontend UI.

\subsection{The App: SwipeRx}
\label{sec:swiperx}

\begin{figure}
    \centering
    \includegraphics[width=0.4\linewidth]{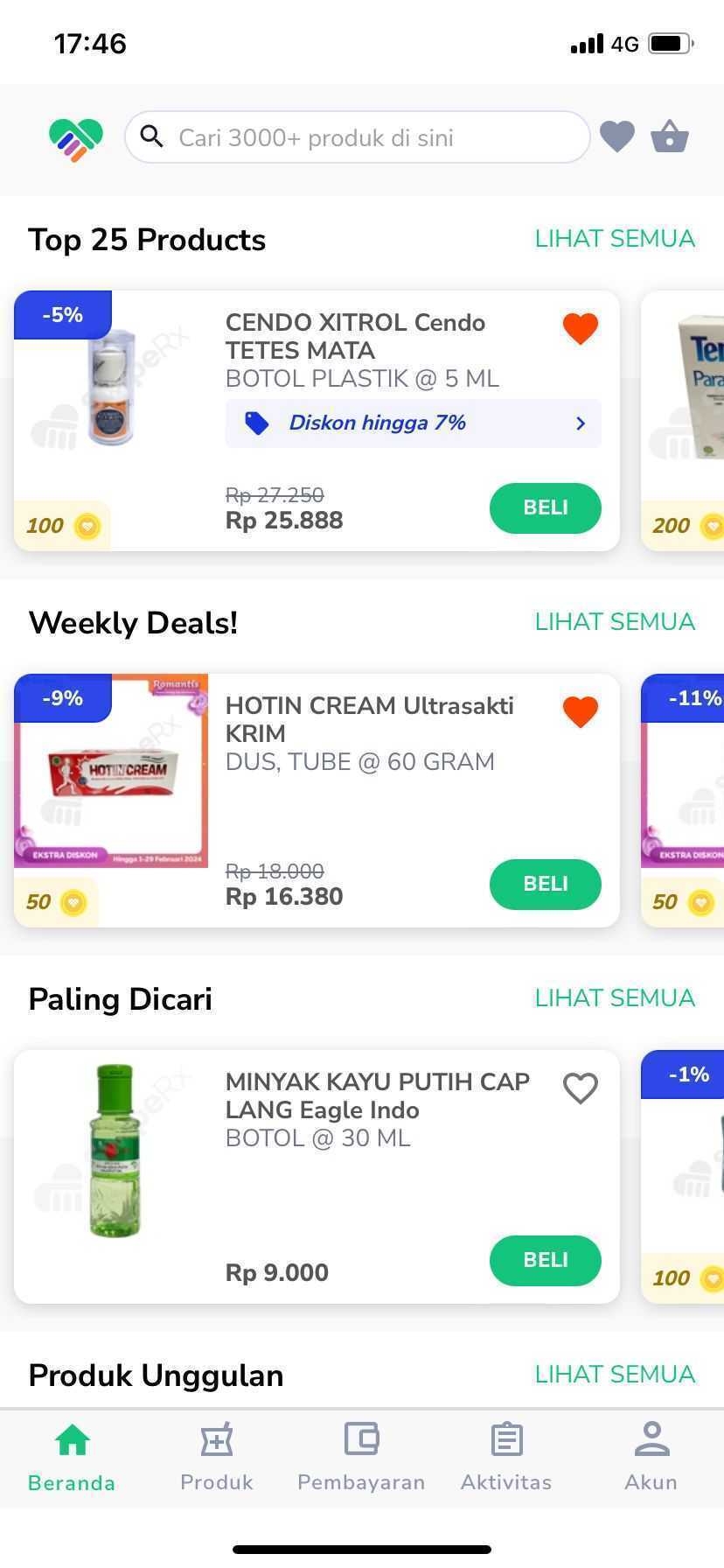} \hspace{0.5cm}
    \includegraphics[width=0.4\linewidth]{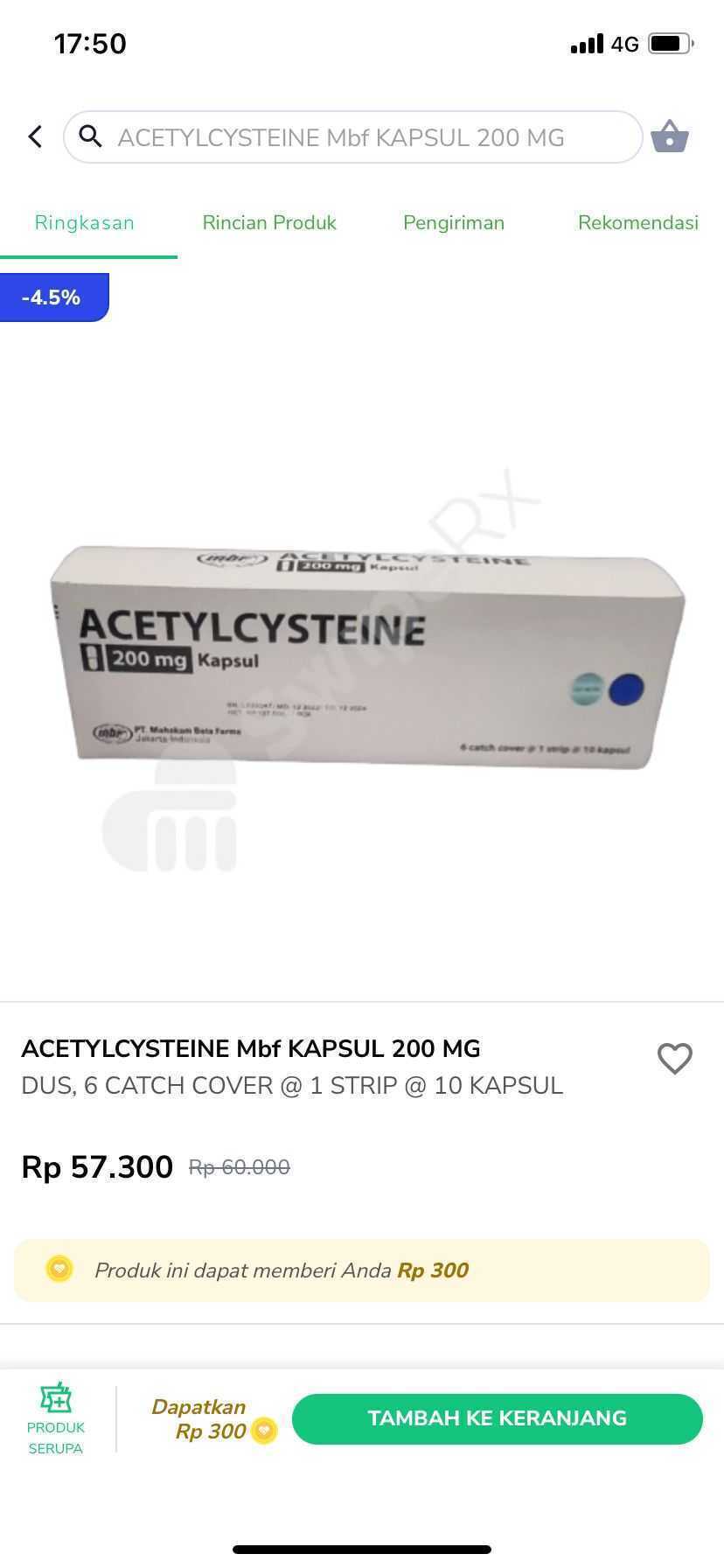}
    \caption{Screenshots of the SwipeRx application.}    \label{fig:screenshots}
    \Description{Screenshots of the SwipeRx application e-commerce section}
\end{figure}

SwipeRx is an all-in-one app for pharmacies in Southeast Asia. It connects over 235,000 professionals across 45,000 pharmacies in Indonesia, the Philippines, Malaysia, Thailand, Cambodia, and Vietnam, providing them with online education, centralized purchasing, logistics and financing, news, drug directory, and adverse event reporting, among other services. Figure \ref{fig:screenshots} shows two screenshots of the app. SwipeRx's community-driven B2B e-commerce model uses medication demand data and bulk purchases to enable small, pharmacy-owned pharmacies (the majority in Southeast Asia) to remain well-stocked by offering medicines at competitive prices and financing mechanisms.

\section{Methodology}

After integrating with the AI platform, the first interventions for SwipeRx users have been adaptive, personalized item (a pharmaceutical product in this context) pair recommendations. These aim to increase the basket size by promoting products the user has seldom or never ordered through the app. Leveraging opportunity costs can have positive spillover effects that improve the availability of relevant supplies in pharmacies. Different recommendation strategies, such as demand forecast-based ones to prevent stock-outs, as well as interventions targeting specific public health concerns through other app functionalities, will be run in the future.

\subsection{General Intervention Setup}

The interventions consist of personalized item pair recommendations (see Section \ref{sec:pair-recommendation}) that can be sent weekly through in-app messages with the text (in Bahasa, the local language): {\textit{Pharmacies in your area typically purchase A and B. Click here to order now!}} If the user is inactive when the in-app message is scheduled, they will find it on their homepage when reopening the app. Messages expire after a week to prevent their accumulation. Users targeted are those associated with pharmacies in Indonesia with only one or two SwipeRx users, with the app language set to Bahasa, which have connected on average at least every week in the last two months and at least once in the previous 40 days. The top 20\% spenders were excluded. The total pharmacy expenditure on the six days following the recommendation was the objective metric used as a reward for the sequential decision-making algorithm (see Section \ref{sec:bandits}) that weekly assigns each user either to treatment (i.e., the message is sent) or control (i.e., the message is not sent). Section \ref{sec:experiments} describes the particularities distinguishing the two concluded experiments run so far with this intervention.

\subsection{Contextual Bandits}
\label{sec:bandits}

RL~\cite{sutton_reinforcement_2018} is the machine learning paradigm best suited to make sequential decisions and, thus, for adaptive intervention delivery. The full Markov Decision Process (MDP)  will be required as the underlying model for interventions where the reconciliation of short and long-term goals and the optimization of a sequence of interventions is critical, which is the case for complex inventory management use cases~\cite{Liu2020, Yu2019, Ie2019, Liu2019, Lillicrap2019, Chen2020}. Multi-Armed Bandits (MABs) are, however, an adequate framework to deal with the interventions described in this paper~\cite{lattimore_bandit_2020, li_contextual-bandit_2010, chu_contextual_2011, agrawal_thompson_2013}, with non-linear models and other extensions providing additional representational complexity to drive the bandit towards regret minimization through more comprehensive personalization~\cite{riquelme_deep_2018, zhou_neural_2020, xu_neural_2022, nabati_online_2021, fedus_revisiting_2020, duran-martin_efficient_2022, smith_application_1962, Daum2015}. The algorithm of choice for this intervention was a Gauss-Gamma linear bandit using Thompson sampling.

Contextless or linear bandits~\cite{li_contextual-bandit_2010, chu_contextual_2011} are a natural setup for adaptive experimentation, allowing for the introduction of statistical power constraints to guide the decision process~\cite{Burtini2015, Yao2021, Dwivedi2022, Xiang2022}. Fully randomized control trials (RCTs), however, remain the golden standard for experimentation 
in both their single- and multiple-assignment designs~\cite{klasnja_microrandomized_2015,qian_microrandomized_2022, Zhang2022, Burtini2015, Yao2021, Dwivedi2022, Xiang2022, Imbens2015}. We can consider our experiments to be combining elements of both. The adaptive intervention can be considered an adaptive experiment, particularly as they were run for a relatively short period and with a linear model on only a handful of basic metrics as context with the primary goal of facilitating knowledge extraction. However, a group of the targeted users was always left out altogether of the adaptive interventions to be used as \textit{pure control} of an RCT intended to discern the impact of the adaptive intervention.

\subsection{Item Pair Recommendation}
\label{sec:pair-recommendation}

We use a rule-based pair recommendation algorithm that considers products typically purchased together, ranks the pairs by revenue, and retains the first 100 pairs, and of them, only those pairs that are currently in stock, creating a list of candidates. When determining what to recommend to a specific user, the algorithm selects from this list of candidate pairs the one that includes one product the user frequently buys and another the user has either not purchased before or has purchased infrequently. Frequent or infrequent here is determined by the average of each user's purchasing history for that item. Priority is given to recommending products the user has not purchased previously, else to those with a larger discrepancy in relative frequencies.

\subsection{Impact Analysis}

The effectiveness of the recommendations is assessed as a combination of three different perspectives.

\subsubsection{RCT angle}

We compare participants' average expenditure (among other engagement metrics) in the adaptive intervention with those in pure control by performing t-tests comparing daily and accumulated (since the beginning of the experiment) expenditure values and analyzing the evolution of significance, effect as given by Cohen's d, and statistical power throughout the experiment. We also use logit regressions on the accumulated expenditures and linear mixed models (LMMs) on the longitudinal expenditure values to estimate the effect of the adaptive intervention with baseline and other covariate adjustments. We probe for heterogeneous effects through a combination of stratified t-tests and estimation of different effects through LMMs.

\subsubsection{Bandit angle}

Within the adaptive intervention, we look mainly into the evolution of the fraction of users assigned to each arm (treatment or control) as dictated by the objective metric observed for these so far. We compute the sensitivity to contextual metrics as the Jacobian of the arm probability based on Thompson sampling approximation soft-thresholded using half the sample's standard deviation and averaged across the participants to understand heterogeneous effects. We plot the t-distributed stochastic neighbor embedding (t-sne) visualization~\cite{hinton_stochastic_2002, maaten_visualizing_2008} using the contextual traits of best arm assignments and the difference in assignment probabilities of the best arm and the second best arm, to gain intuition on whether the context as a whole was well chosen.

\subsubsection{Recommendation success angle}

Finally, we also look into how the user reacted to the recommendation (if they opened it, closed it, or ignored it) and whether the recommended item (the one previously infrequently purchased) was purchased at a later time during the experiment. 

\subsection{Experiments}
\label{sec:experiments}

Two experiments have been completed so far with the item pair recommendations. Besides running over different periods (they have been run consecutively), they used different contexts for the adaptive arm. Sample sizes also differ, as all participants in the adaptive intervention of the first experiment (XP1) were excluded from the second one (XP2). The in-app message is sent at 6 am local time on Monday for both experiments except for the first four weeks of XP1 when they were scheduled for 6 am Tuesday. This change was made mid-experiment to match the day of recommendation with the day of the week when a large part of all purchases and logins happen. 

\subsubsection{First item pair recommendation experiment (XP1)}

Ran for 8 weeks between December 2023 and January 2024, with around 30\% of cohort users assigned to pure control. Context used consisted in the region (one-hot encoded), normalized days since the last nudge, days with purchase orders and expenditure in the last 90 days.

\begin{figure*}[ht!]
  \centering
  \includegraphics[width=0.8\linewidth]{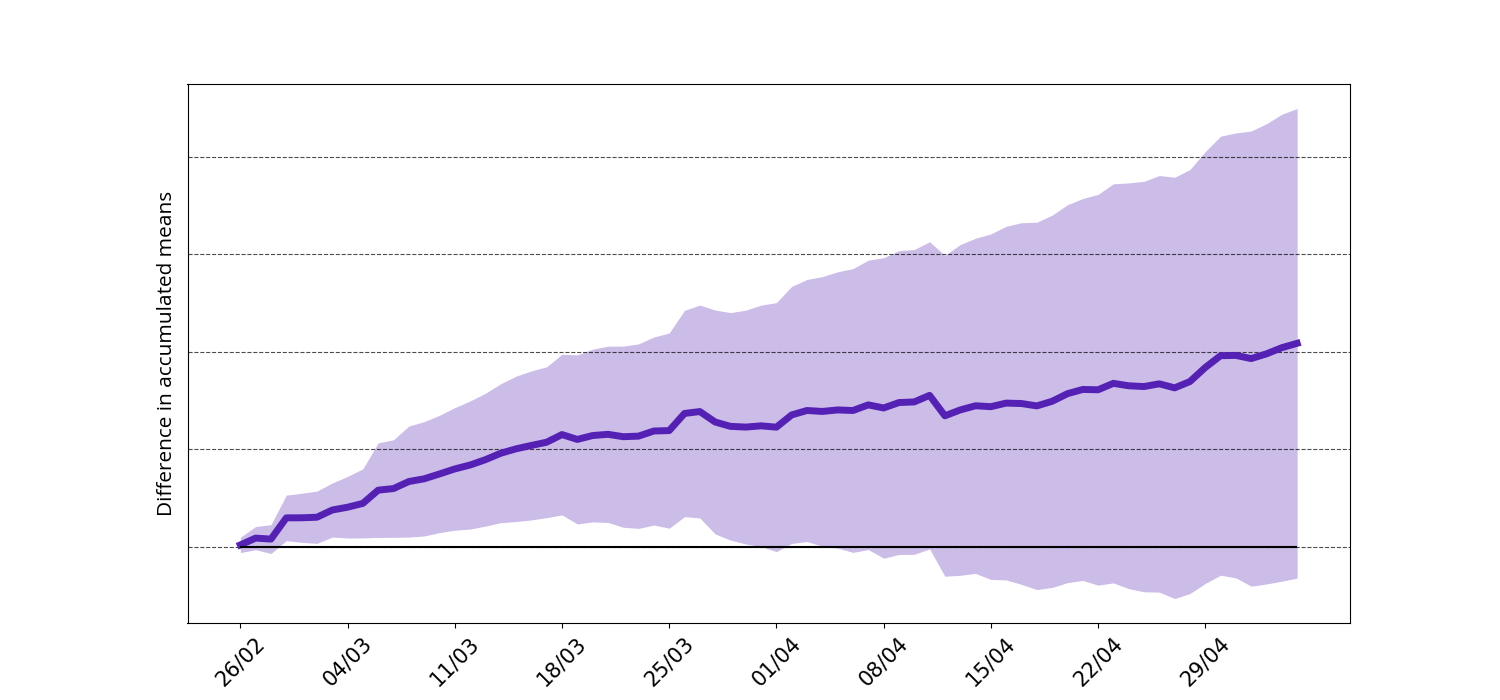}
  \caption{Difference between mean accumulated daily expenditure between users in the adaptive arm vs. control for all users in XP2. Confidence intervals (90 \%) are shaded. The black horizontal line is drawn across 0, and vertical lines represent the intervention period's beginning and end.}
  \label{fig:rct-acc}
  \Description{Plot showing the growing accumulated difference in mean expenditure as experiment XP2 unfolds with a long period of over a month where the increase is statistically significant.}
\end{figure*}

\subsubsection{Second item pair recommendation experiment (XP2)}

Ran for 10 weeks between February and April 2024, with around 40\% of participants assigned to pure control. Context used consisted on average days between logins in the last 60 days, days since first login, expenditure in the last month, days with orders in the last 30 days, normalized days since last nudge, nudges opened in the last 2 weeks, and time spent using the app in the last 30 days.

\section{Results}
\label{sec:results}

The experiments consistently indicate a small positive impact on pharmacy expenditure, stabilizing once the users are habituated. There is also evidence of delayed effects, i.e., recommendations not acted on immediately but purchased later. This section summarizes our main findings. The complete analysis is not included due to length limitations but will be published elsewhere \cite{work-in-progress}. Table \ref{tab:results} collects a few impact metrics across the two experiments.

\begin{table}
  \caption{Impact metrics across experiments. T-test metrics refer to comparing the daily accumulated expenditure (since the beginning of the experiment) between the adaptive intervention and pure control for significant days only. LMM refers to the estimated parameters of a linear mixed effects model on weekly values where all parameters are significant. Bandit metrics indicate the average percentage of the participants in the adaptive intervention assigned to treatment across the experiment and the fraction of decision points where most users (> 50\%) were sent the nudge. Successful recommendations are those that result in purchases at a later time during the experiment of the pair's item ordered infrequently by the user. A significance level of 90\% is used, and - indicates nonsignificant results.}
  \label{tab:results}
  \begin{tabular}{p{5cm} p{1cm} p{1cm}}
    \toprule
    &XP1&XP2\\
    \midrule
     T-test: days with significant effect  & 0\% & 57\% \\
    T-test: largest effect & - & 0.12 \\
    T-test: largest statistical power & - & 0.65 \\
    T-test: average effect & - & 0.10 \\
    T-test: average statistical power & - & 0.55 \\   
    LMM: adaptive intervention arm & 5.97 & -  \\
    LMM:  nudged that week   & 15.99 & 14.17 \\
    LMM: baseline expenditure & 0.11 & 0.10  \\
    Bandit: assigned to nudge & 26\% & 70\% \\
    Bandit: majority assigned to nudge & 1/8 & 8/9 \\
    Successful recommendations & 18.2\% & 22.9\% \\
\end{tabular}
\end{table}

While the t-test accumulated expenditure throughout XP1 was not significant at any point for the required level of confidence, it came close, particularly after the change in the nudge schedule. Some daily values were significantly higher in the adaptive arm and for daily and accumulated values for certain strata, such as pharmacies in Jakarta. The LMM also estimated a significant positive impact of being in the adaptive arm despite the bandit learning to assign decreasing numbers of users to treatment. The sensitivity analysis suggests the context helped discriminate which users would benefit from the recommendation, with the days since the last nudge preventing fatigue effects and the purchasing frequency having a large negative impact on the probability of assigning to treatment. All of this, together with the fact that this experiment ran over the atypical Christmas period, encouraged an experimental design for XP2 that mirrored that of XP1, with the scheduling on Monday since the beginning, with only a modified context.

For XP2, despite the smaller sample sizes used, we see a growing impact in accumulated values that reaches significance after a week and remains so for 40 days, as depicted in Figure \ref{fig:rct-acc}. Opposite to XP1, the bandit in XP2 learns to nudge most users after three assignments. This could be related to the allocation by chance at that decision point of a large fraction of the top spenders to treatment, skewing its learning process. In turn, this appears to hinder the bandit's capacity to adequately learn how to use its context, as indicated by the sensitivity analysis, where, for example, most traits were negligible and the impact of days since the last nudge the opposite of XP1 and is reasonable to expect.
The large correlation between being in the adaptive arm and being nudged close to every week also explains why, for XP2, the LMM only estimates the parameter associated with the latter as significant (opposite to XP1).

The faction of successful recommendations is very similar and relatively large in both XP1 and XP2. 
The LMM baseline estimated effects are also consistent in both cases. Exploration of heterogeneous effects through stratified analysis, LMM effect estimation, and sensitivity analysis indicate areas that should be explored in more detail, such as regional effects and the impact of the user's typical purchasing frequency, as the pairs appear to be particularly useful in nudging both very frequent purchasers and extremely infrequent purchasers.

\section{Summary and Conclusions}

We developed a framework to enhance digital tools for pharmacies and healthcare providers with RL-based behavioral nudging. Applied to SwipeRx, a leading app for Southeast Asian pharmacists, our approach is illustrated through interventions aiming to increase basket size by helping pharmacists discover new products. The positive impact observed in our experiments underscores the potential of this approach to support and improve pharmacy services.

\begin{acks}
The authors want to thank Susan Murphy for insightful discussions. This work was supported, in whole or in part, by the Bill \& Melinda Gates Foundation INV-060956. Under the grant conditions of the Foundation, a Creative Commons Attribution 4.0 Generic License has been assigned to the Author Accepted Manuscript version that might arise from this submission.
\end{acks}

\bibliographystyle{ACM-Reference-Format}
\bibliography{main}




\end{document}